\documentclass[pdflatex,sn-vancouver-num]{sn-jnl}

\usepackage{graphicx}%
\usepackage{multirow}%
\usepackage{amsmath,amssymb,amsfonts}%
\usepackage{amsthm}%
\usepackage{mathrsfs}%
\usepackage[title]{appendix}%
\usepackage{xcolor}%
\usepackage{textcomp}%
\usepackage{manyfoot}%
\usepackage{booktabs}%
\usepackage{algorithm}%
\usepackage{algorithmicx}%
\usepackage{algpseudocode}%
\usepackage{listings}%
\usepackage{booktabs}
\usepackage{array} 

\theoremstyle{thmstyleone}%
\theoremstyle{thmstyletwo}%

\theoremstyle{thmstylethree}%

\begin{document}

\title[Article Title]{MR-UIE: Multi-Perspective Reasoning with Reinforcement Learning for Universal Information Extraction}

\author[1]{\fnm{Zhongqiu} \sur{Li}}\email{lizq48@chinatelecom.cn}
\equalcont{These authors contributed equally to this work.}
\author[1]{\fnm{Shiquan} \sur{Wang}}\email{wangsq23@chinatelecom.cn}
\equalcont{These authors contributed equally to this work.}
\author[1]{\fnm{Ruiyu} \sur{Fang}}\email{fangry@chinatelecom.cn}
\equalcont{These authors contributed equally to this work.}

\author[1]{\fnm{Mengjiao} \sur{Bao}}\email{baomj@chinatelecom.cn}
\author[1,2]{\fnm{Zhenhe} \sur{Wu}}\email{wuxs97@163.com}
\author[1]{\fnm{Shuangyong} \sur{Song}}\email{songshy@chinatelecom.cn}
\author*[1]{\fnm{Yongxiang} \sur{Li}}\email{liyx25@chinatelecom.cn}
\author*[1]{\fnm{Zhongjiang} \sur{He}}\email{hezj@chinatelecom.cn}


\affil[1]{\orgdiv{Institute of Artificial Intelligence (TeleAI)}, \orgname{China Telecom Corp Ltd}, \country{Xicheng District, Beijing, China}}
\affil[2]{\orgdiv{BeiHang University}, \country{HaiDian District, Beijing, China}}

\abstract{
Large language models (LLMs) demonstrate robust capabilities across diverse research domains, however, their performance in universal information extraction (UIE) remains insufficient, especially when trackling structured output scenarios that involve complex schema descriptions and requiring multi-step reasoning. While existing approaches enhance the performance of LLMs through in-context learning and instruction tuning, significant limitations nonetheless persist. To enhance the model's generalization ability, we propose integrating reinforcement learning (RL) with multi-perspective reasoning for information extraction (IE) tasks. Our work transitions LLMs from passive extractors to active reasoners, enabling them to understand not only what to extract but also how to reason.
Experiments conducted on multiple IE benchmarks demonstrate that MR-UIE consistently elevates extraction accuracy across domains and surpasses state-of-the-art methods on several datasets.Furthermore, incorporating multi-perspective reasoning into RL notably enhances generalization in complex IE tasks, underscoring the critical role of reasoning in challenging scenarios.
}




\maketitle

\section{Introduction}\label{sec1}

Information extraction (IE) is a fundamental task in natural language processing (NLP), which encompasses a wide range of subtasks such as Named Entity Recognition (NER), Relation Extraction (RE), and Event Extraction (EE)
\cite{li2023evaluating,ma2023large,xu2023large,li2023codeie}. Traditionally, these tasks have been addressed by specialized models trained in task-specific datasets. However, the fragmentation of tasks and schemas has hindered the development of generalizable and scalable IE tasks.

To address this limitation, recent research has focused on universal information extraction (UIE), which aims to model all IE tasks within a universal framework. A seminal work in this direction is proposed by Lu et al., which introduced a structured generation paradigm that encodes diverse IE tasks into a common semantic representation\cite{DBLP:conf/acl/0001LDXLHSW22}. Building on this, InstructUIE\cite{DBLP:journals/corr/abs-2304-08085} extended the idea by incorporating multi-task instruction tuning, enabling models to generalize across tasks via natural language instructions.


\textcolor{teal}{}
With the emergence of powerful  LLMs\citep{he2024telechat,li2024tele,wang2024telechat,li202452b,wang2025technical}, significant advancements have been made across long-standing NLP tasks such as text classification\citep{song2017intension,song2015classifying,song2020sentiment,zhao2022tosa,zhang2022multi}, intent recognition\citep{xu2023improving,pang2022mfdg}, entity linking\citep{miao2016automatic,song2010spatio,song2016linking,song2018linking}, and beyond. Inspired by their robust performance and adaptability, researchers have explored their potential for information extraction through prompting and in-context learning learning\citep{ye2023comprehensive,chen2023robust}. For example, CodeIE demonstrated that code generation models can serve as strong few-shot IE extractors by using structured code-like commands\citep{DBLP:conf/acl/LiSTYWHQ23}. Similarly, ChatUIE and YAYI-UIE explored chat-based and instruction-tuned LLMs for UIE, emphasizing the importance of human-like interaction and schema-guided generation\cite{DBLP:conf/coling/XuSZZ24,DBLP:journals/corr/abs-2312-15548}.
Meanwhile, \cite{dai2025r1} proposes the R1-RE framework, which incorporates the human annotation reasoning process of "recognition–comparison–reasoning–verification" into relation extraction, guiding the model to learn human-like reasoning patterns through a multi-stage reward mechanism. In contrast, \cite{mo2024c} enhances the model’s reasoning capability from a data construction perspective by integrating positive and negative examples and filtering hard negatives to help the model correct its reasoning errors. Both approaches offer effective strategies for reasoning-driven information extraction, from the perspectives of human reasoning simulation and data augmentation, respectively.

\begin{figure*}[htbp]
    \centering
    \includegraphics[width=\linewidth, scale=5]{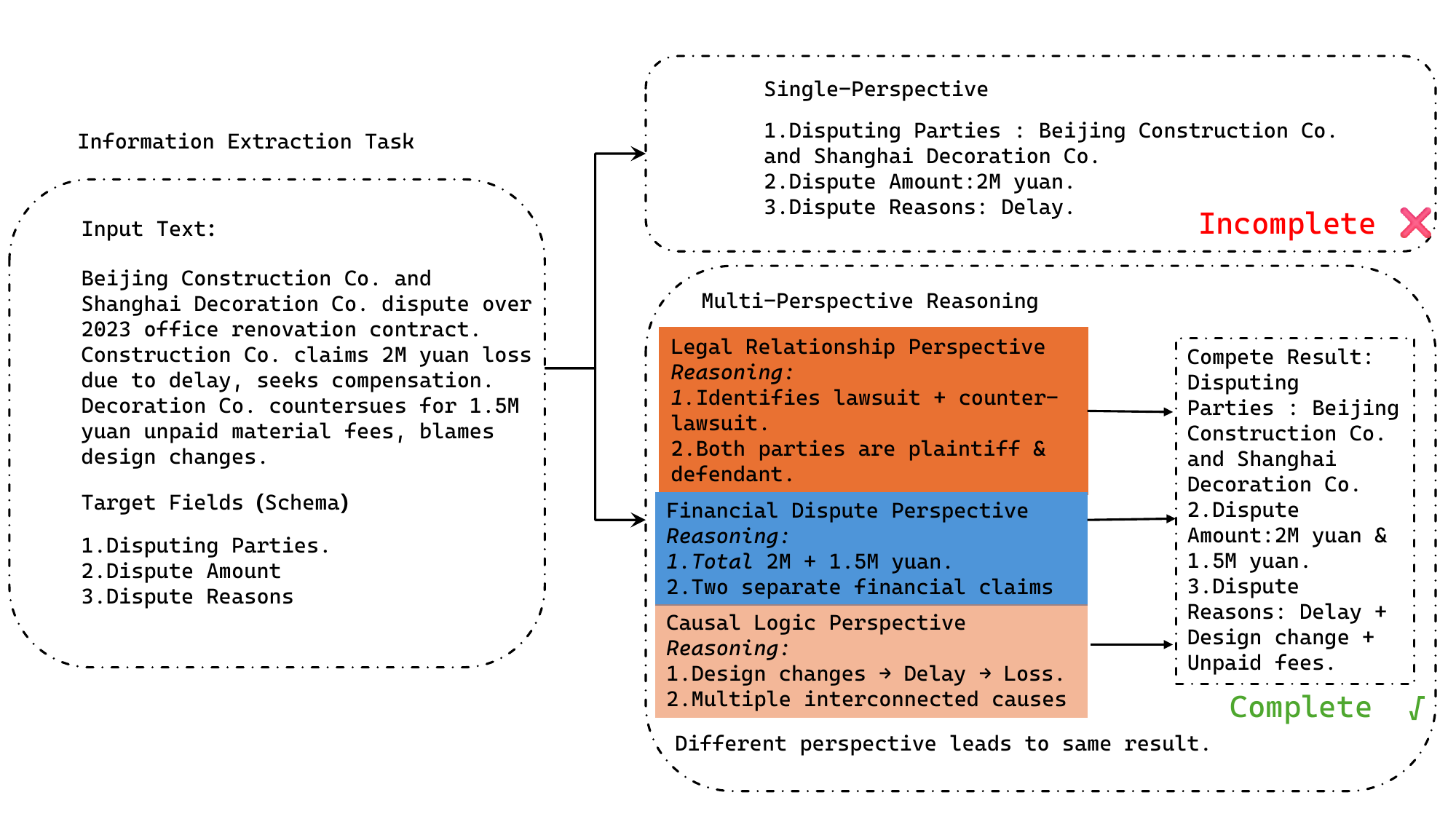}
    \caption{The case of multi-perspective reasoning. The multiple analytical viewpoints generates richer contextual representations during the extraction process, thereby expanding the variability and coverage of processed information.}
    \label{fig:example}
\end{figure*}

Despite these advances, the current LLM-based UIE model still faces several challenges. first, existing methods still struggle to address complex information extraction (IE) scenarios, which involve unstructured texts replete with ambiguity, implicit semantic relationships, and context-dependent information. The figure~\ref{fig:example}~ilustrate a case with multi-event news articles where causal links between events are not explicitly stated. In such cases, extracting target entities, relations, or events cannot be achieved through simple pattern matching or single-step classification. Instead, it demands iterative reasoning: first disambiguating ambiguous references (e.g., resolving pronouns or coreferential expressions), then inferring implicit connections (e.g., deducing causal relationships between events based on contextual cues), and finally validating the consistency of extracted information against the broader context. Second, reasoning models like \textit{Deepseek-R1}~\citep{guo2025deepseek} and \textit{OpenAI-o1}~\citep{openai2024gpt4technicalreport} have shown that chain-of-thought (CoT) reasoning can enhance  performance in complex reasoning tasks\citep{zhang2024lemur,xing2025llmsr,xiong2025tablereasoner,xiong2025teleai,zhao2025enhancing,zhang2025t2r,wu2025ucs,wu2025mr,DBLP:conf/semweb/LiWLHFZZLLS24}. However, Little research has explored how information extraction can leverage thinking and reasoning capabilities to improve its performance.

While CoT reasoning has shown promise in NLP tasks, existing approaches typically employ uniform reasoning patterns that can overlook critical information dimensions. We propose that complex IE scenarios require multi-perspective chain-of-thought reasoning, where different analytical angles—such as entity relationships, temporal logic, and semantic roles—work synergistically to ensure comprehensive extraction. This approach is particularly valuable for intricate documents where information spans multiple semantic layers. 

Our approach is inspired by recent work such as DILUIE\cite{DBLP:journals/nca/GuoGZ24}, which emphasizes the importance of diverse reasoning paths, and ADELIE\cite{DBLP:conf/emnlp/Qi0W00L24}, which aligns LLMs with human preferences for IE tasks. 
However, distinct from these methods, we explicitly formulate the reasoning process as a structured multi-path decision problem. Inspired by the work of Table-R1\citep{wu2025table}, we introduce a unified representation to bridge heterogeneous IE tasks and use reinforcement learning to align LLMs with high-quality reasoning strategies.Specifically, our method involves the following steps:
\begin{enumerate}
    \item \textbf{Unified Schema Representation}: We propose a task-agnostic schema abstraction that supports entities, relations, and events in a unified manner. This facilitates schema-conditioned generation and cross-task generalization.
    
    \item \textbf{Multi-perspective Reasoning Path Construction}: For each input, we prompt the model to generate multiple reasoning trajectories based on diverse prompting strategies (e.g., schema-driven, type-first, evidence-first), capturing different inferential perspectives.
    
    \item \textbf{Adaptive Reasoning Model Training}: We guide the model to conduct structured inference along each path and generate interpretable intermediate steps. During training, we use path-aware supervision to help the model learn reasoning procedures beyond final answers.
    
    \item \textbf{Reinforcement Learning for Reasoning Optimization}: Finally, we design a composite reward function that evaluates the correctness, faithfulness, and efficiency of reasoning traces. We apply reinforcement learning to help the model internalize the most effective reasoning patterns over time.
\end{enumerate}

\textbf{Our main contributions are:}

\begin{itemize}
    \item We propose a novel UIE framework that integrates structured inference and reinforcement learning to explicitly model the reasoning process.
    
    \item We design a multi-perspective reasoning path generation mechanism that captures diverse and complementary inference styles.
    
    \item We demonstrate that reasoning-aware training improves generalization and interpretability across multiple IE benchmarks.
\end{itemize}
By treating LLMs as \textbf{adaptive reasoners} rather than static predictors, our framework encourages them to learn not only what to extract, but how to reason effectively in complex extraction scenarios.

\section{Related Work}\label{sec2} 
\subsection{Universal Information Extraction}
\label{sec:UIE}
Information Extraction (IE), a fundamental task in natural language processing, aims to identify and structure key information from unstructured text\citep{lample2016neural,zheng2017joint,song2023towards,wang2024enhancing,lin2019sequence}. Traditional IE approaches are primarily categorized into two types based on their constraints regarding information \textbf{schema}---predefined structured frameworks: Closed IE and Open IE\cite{mao2025classifier,shao2025ai,an2025ai}. Closed IE requires extracted information to strictly conform to a predefined schema, encompassing several core subtasks: Named Entity Recognition (NER) identifies and classifies entities into predefined categories; Relation Classification (RC) determines predefined relationship types between mentioned entities; Relation Extraction (RE) jointly extracts entities and their relations; Event Detection (ED) identifies event triggers and classifies them into predefined types; Event Argument Extraction (EAE) extracts role-specific arguments for events; Event Extraction (EE) performs end-to-end event and argument identification; and Event Relation Extraction (ERE) extracts predefined relationships between events (e.g., temporal, causal). Conversely, Open IE extracts open-domain $n$-ary relation tuples (e.g., \textit{(subject; relation; object)}) without schema constraints, offering greater flexibility at the cost of structural consistency. To unify these diverse paradigms, \textbf{Universal Information Extraction (UIE)} has emerged, utilizing shared representations and unified architectures to handle multiple IE tasks~\citep{DBLP:conf/acl/0001LDXLHSW22}. Our work focuses specifically on \textbf{Closed UIE}, investigating enhanced modeling approaches for schema-constrained extraction within universal frameworks through alignment optimization.

\subsection{LLMs based Universal Information Extraction}
\label{sec:LLM-UIE}

Large language models have revolutionized UIE by enabling unified frameworks that overcome the schema constraints of traditional Closed IE systems. Building on the foundation of schema-constrained extraction established in Section \ref{sec:UIE}, recent approaches leverage LLMs' emergent capabilities to handle diverse IE tasks through shared architectures and representations. This paradigm shift is driven by three key innovations in alignment methodologies.

Instruction engineering has emerged as a critical technique for schema alignment, particularly for Closed UIE tasks requiring strict adherence to predefined structures. \textit{InstructUIE}~\citep{DBLP:journals/corr/abs-2304-08085} leverages expert-crafted templates to explicitly model task dependencies, achieving 12\% F1 improvement in few-shot scenarios while outperforming specialized models on 85\% of IE tasks. \textit{CodeIE}~\citep{DBLP:conf/acl/LiSTYWHQ23} transforms this approach by reformulating structured outputs as code, enabling precise schema compliance through code-generation LLMs that outperform natural-language LLMs in few-shot settings. For dynamic schema adaptation, \textit{GOLLIE}~\citep{DBLP:conf/iclr/SainzGALRA24} incorporates human annotation guidelines as structured priors, enabling robust zero-shot generalization to new domains with only 3-5 examples via dynamic constraint decoding. These techniques are enhanced by the contrastive learning framework of \textit{C-ICL}~\citep{DBLP:conf/emnlp/MoLYWZWL24}, which improves boundary robustness through hard negative examples and code-style templates. Moreover,\textit{DILUIE}~\citep{DBLP:journals/nca/GuoGZ24} operationalizes a demonstration-diversity principle that expands coverage of semantic variations and edge cases.

Data-centric alignment methods have significantly advanced schema-conditioned extraction through large-scale, high-quality corpora. \textit{IE-Pile}~\citep{DBLP:conf/acl/GuiYYZSLC24} constructs the first massive schema-conditioned instruction corpus (IE-Instruct), enabling state-of-the-art performance across multiple Closed IE benchmarks through targeted instruction tuning. Scaling this approach, \textit{ADELIE}~\citep{DBLP:conf/emnlp/Qi0W00L24} expands to 83K+ instructions and employs Direct Preference Optimization (DPO) for complex schema alignment, featuring dynamic schema adjustments like class-name substitution that enhance zero-shot transferability. These datasets specifically address the schema constraints inherent in Closed UIE by explicitly encoding structural requirements during model alignment.
\begin{figure*}[htbp]
    \centering
    \includegraphics[width=\linewidth, scale=5]{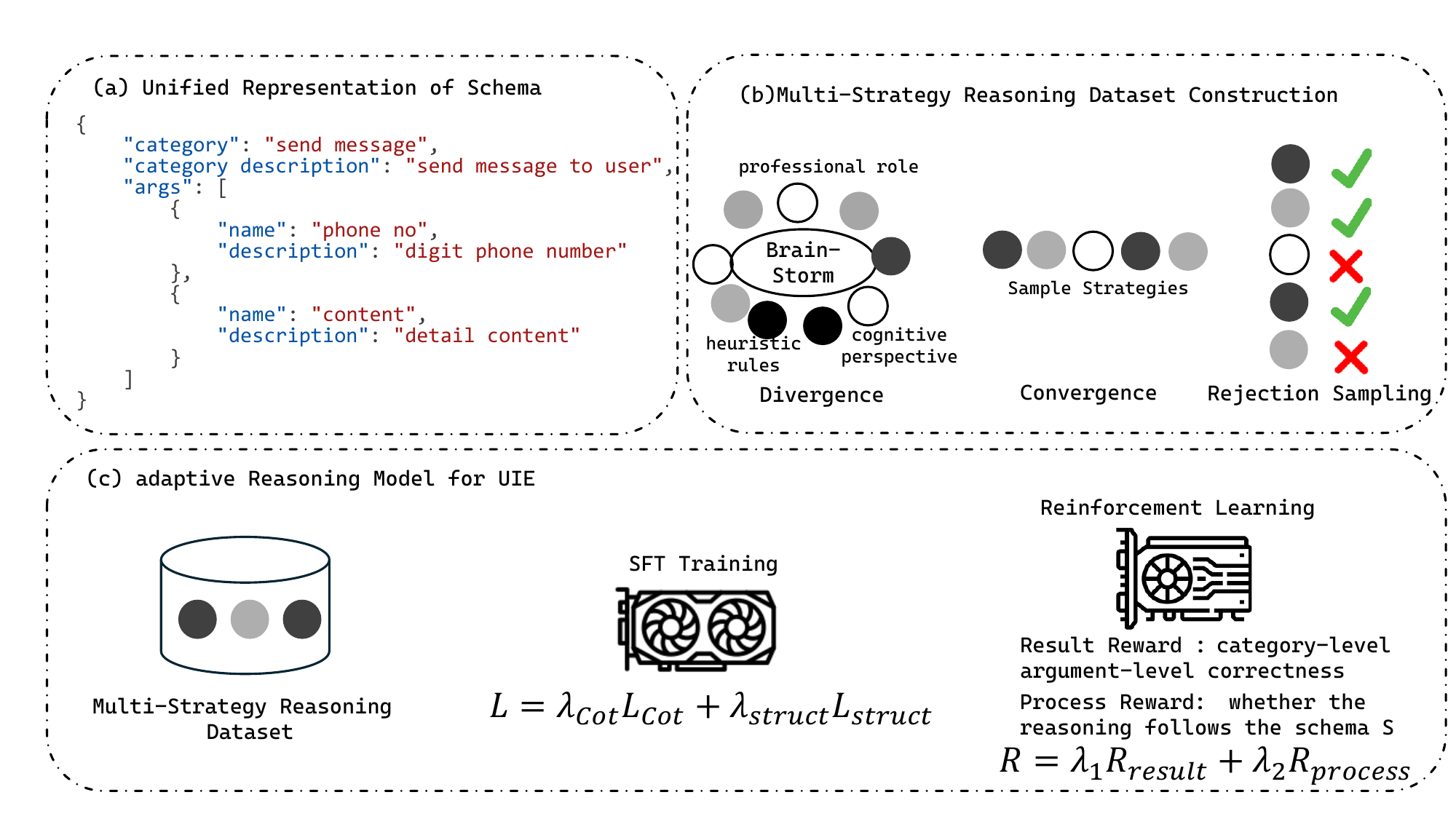}
    \caption{The overview of MR-UIE. We propose multi-perspective reasoning with: (a) Unified Schema Representation (hierarchical JSON with \(c_i\), \(a_i\), \(d_i\)) for universal Input-Output Constraints; (b) Multi-Perspective Generation (divergence: \(3N\) strategies across ``cognitive'', ``role'', ``heuristic dimensions'' \(\rightarrow\) convergence: paradigm-based selection \(\rightarrow\) rejection sampling); (c) Adaptive Reasoning (SFT + RL alignment) with multi-grained rewards (\(R_{\text{result}}\), \(R_{\text{process}}\)) for optimal path selection. Demonstrates reinforcement-guided universal information extraction.}
    \label{fig:framework}
\end{figure*}
Architectural innovations have optimized LLMs for structured output generation essential for schema-compliant extraction. \textit{ChatUIE}~\citep{DBLP:conf/coling/XuSZZ24} integrates reinforcement learning with generation constraints in conversational models, achieving performance surpassing fully supervised systems while maintaining schema integrity. \textit{YAYI-UIE}~\citep{DBLP:journals/corr/abs-2312-15548} refines this paradigm through two-stage, chat-augmented tuning for bilingual schema adaptation, improving cross-lingual transfer in predefined schema environments. Emerging retrieval-augmented approaches like \textit{RUIE}~\citep{DBLP:conf/coling/LiaoDH025} demonstrate potential for contextualizing extraction within schema constraints through dynamic knowledge grounding.

Current LLM-based UIE systems excel in schema-constrained environments through code-formalized outputs, contrastive boundary learning, and dynamic schema adaptation. Future advancements will focus on optimizing complex schema reasoning, cost-efficient alignment for domain-specific schemas, and retrieval-augmented architectures for knowledge-intensive Closed IE applications. These developments directly address the core challenge of our work: enhancing schema-constrained extraction within universal frameworks through alignment optimization.

\textcolor{teal}{}Despite the progress achieved by instruction tuning, schema-conditioned corpora, and model-level enhancements, existing UIE systems often struggle with:
(1) reasoning under complex schema constraints;
(2) adapting to task- or domain-specific variations without explicit demonstration;
(3) optimizing extraction alignment beyond local token- or sequence-level objectives.
To address these limitations, we propose a unified framework that integrates multi-perspective reasoning, adaptive instruction alignment, and reinforcement-based extraction path optimization.

\section{Method}\label{sec3}
We propose a multi-perspective reasoning for UIE model, which enables large language models to perform universal IE tasks with adaptive reasoning paths. As shown in Figure~\ref{fig:framework}, our approach consists of four core components.
\subsection{Unified Representation of Schema}\label{UR-schema}

Drawing on function call semantics, we establish a canonical schema representation that transforms task-specific schemas into a unified JSON structure. This framework encodes extraction targets through hierarchical key-value pairs with three core components: class identifiers (\textit{c\textsubscript{i}}), property-defining arguments (\textit{a\textsubscript{i}}), and semantic descriptors (\textit{d\textsubscript{i}}). The descriptor component enables precise label alignment while maintaining structural consistency across diverse information extraction tasks through embedded domain knowledge.

This representation enables schema-constrained decoding during model training, where output structures are dynamically validated against the canonical grammar. During inference, unseen task schemas are projected into this unified space through descriptor embeddings, eliminating task-specific adapters and enabling zero-shot generalization without structural modifications.
\subsection{Multi-perspective Reasoning Dataset Construction}
\label{sec:reasoning-dataset}

Most of the UIE models are typically trained on instruction-following datasets that impose a fixed prompting strategy.  
This rigidity hinders their ability to tailor inference paths to the intricate interplay between textual complexity and schema characteristics.  
Drawing inspiration from the human cognitive cycle of ``multi-perspective brainstorming $\rightarrow$ multi-solution generation $\rightarrow$ convergence,'' we construct a multi-perspective reasoning dataset that elicits and distills a compact yet diverse set of core reasoning patterns aligned with each input pair.

\textbf{Divergence Phase.}
Given an input text $x$ and its associated schema $S$, we leverage LLMs as a strategy generator.  
The prompt enumerates three high-level analytical dimensions---(1) cognitive perspective, (2) professional role, and (3) heuristic rules---and instructs the model to independently produce $N$ fine-grained thinking strategies for each dimension, yielding $3N$ raw strategies in total.

\textbf{Convergence Phase.}
To consolidate these raw strategies, we first define keyword paradigms, each comprising a curated list of indicative keywords.  
Each strategy is assigned to the paradigm whose keywords exhibit the highest co-occurrence with the strategy text.  
Within every paradigm cluster, we compute TF-IDF-based embeddings and measure pairwise cosine similarities.  
We then identify two representative strategies: the one with the lowest average similarity to all other strategies (maximally unique) and the one with the highest average similarity (maximally generic), thereby balancing distinctiveness and representativeness.  
From this paradigm-refined candidate pool, we uniformly sample $P$ strategies $\mathcal{T} = \{\tau^1, \dots, \tau^P\}$ as our core set.  
For each $\tau^p$, we prompt \textsc{DeepSeek-R1} to produce a step-by-step chain-of-thought (\textsc{CoT}) rationale together with a structured prediction $y$, we formalize
the ICL process as:
\begin{equation}
\mathcal{M}(x_i, S, \tau^p) \rightarrow \{CoT_i^p,y_i^p\}
\end{equation}

\textbf{Rejection Sampling.}
Finally, we compare each prediction~$y_{i}$ against the ground-truth label~$y^{*}$.  
Any strategy whose prediction satisfies~$y_{i}\neq y^{*}$ is discarded.  
For every sample, let~$k$ be the number of surviving strategies; by construction~$0\leq k\leq P$.  
We introduce a minimum-threshold parameter~$O$.  
If~$k \geq O$, the sample together with its~$k$ valid strategies enters the reasoning dataset.  
Conversely, if~$k< O$, the sample is withheld and forwarded to reinforcement learning. Algorithm~\ref{alg:build-dataset-v1} presents the detailed construction process of the Multi-perspective Reasoning Dataset.

\begin{algorithm}[t]
\caption{Multi-perspective Reasoning Dataset Construction}\label{alg:build-dataset-v1}
\begin{algorithmic}[1]
\Require {Text $x$, schema $S$, ground-truth $y^*$, generator $G$ (DeepSeek-R1), 
          strategies-per-dimension $N$, paradigm count $M$, sampling size $P$}
\Ensure {Dataset $\mathcal{D}$ containing up to $P$ valid instances $\langle x, S, \tau, \text{CoT}, y \rangle$}
\State $\mathcal{D} \gets \emptyset$
\State \textit{RawStrats} $\gets$ empty list
\For{\textbf{each} analytical dimension $\Delta \in \{\text{cognitive},\,\text{role},\,\text{heuristic}\}$}
    \For{$i = 1$ \textbf{to} $N$}
        \State $\sigma \gets G(\text{prompt}(x, S, \Delta))$ \Comment{Generate fine-grained strategy}
        \State \textit{RawStrats}.append($\sigma$)
    \EndFor
\EndFor
\State \textit{Paradigms} $\gets$ \{keyword list$_1$,\dots,keyword list$_M$\}
\State \textit{Clusters} $\gets$ assign\_to\_paradigms(\textit{RawStrats}, \textit{Paradigms})
\State \textit{Candidates} $\gets$ empty list
\For{\textbf{each} cluster $C$}
    \State $E \gets$ TF-IDF\_embeddings($C$)
    \State $\tau_{\text{unique}} \gets \arg\min_{\sigma \in C}\frac{1}{|C|}\sum_{\sigma' \in C}\cos(E_\sigma, E_{\sigma'})$
    \State $\tau_{\text{generic}} \gets \arg\max_{\sigma \in C}\frac{1}{|C|}\sum_{\sigma' \in C}\cos(E_\sigma, E_{\sigma'})$
    \State \textit{Candidates}.append($\tau_{\text{unique}}, \tau_{\text{generic}}$)
\EndFor
\State \textit{CoreStrats} $\gets$ uniform\_sample(\textit{Candidates}, $P$)
\For{\textbf{each} $\tau \in \textit{CoreStrats}$}
    \State $(\text{CoT}, y) \gets G(\text{prompt}(x, S, \tau))$
    \If{$y = y^*$}
        \State $\mathcal{D} \gets \mathcal{D} \cup \{\langle x, S, \tau, \text{CoT}, y \rangle\}$
    \EndIf
\EndFor
\State \Return $\mathcal{D}$
\end{algorithmic}
\end{algorithm}

\subsection{Adaptive Reasoning Model for UIE}
\label{sec:training}
To enable the model to steadily learn format compliance, multi-perspective reasoning, and adaptive strategy exploration, we developed a three-stage training strategy, progressively evolving a general information extraction (IE) model into one that autonomously selects analysis strategies for reasoning.

\subsubsection{Supervised fine-tuning for base UIE}\label{subsubsec1}
To achieve broad coverage across diverse IE tasks, our primary objective is to develop a base model equipped with UIE capabilities. Our approach leverages the large-scale, bilingual (Chinese and English) IEPILE dataset, which comprehensively encompasses the three core IE tasks (NER, RE and EE). We utilize the dataset as the primary training resource. To align the data with our proposed unified schema, we retain the original input text $x$, the structured schema $S$, and the corresponding ground-truth annotations $y^*$. Subsequently, we apply a rigorous data refinement pipeline, including: \textbf{Prompt-based Transformation}: Converting instances into instruction-following formats. \textbf{Format Restructuring}: Ensuring output adheres strictly to the unified schema specifications. \textbf{Instance Deduplication}: Removing redundant samples to improve data quality and training efficiency. \textbf{Quality Filtering}: Eliminating instances with errors, inconsistencies, or low annotation quality.
This comprehensive pre-processing results in a curated corpus of one million high-quality instances specifically optimized for training a general-purpose IE model.

Following data curation, we employ SFT with an instruction tuning paradigm focused on text-to-structure generation. This fine-tuning phase explicitly teaches the model to interpret natural language instructions based on the predefined universal schema and to generate structured outputs that strictly conform to the specified format constraints. The resulting instruction-tuned model demonstrates robust general-purpose information extraction capabilities. It effectively handles diverse IE sub-tasks within a single, unified framework, consistently producing structured representations governed by the universal schema.
\subsubsection{SFT with Multi-perspective Reasoning}\label{subsubsec2} 
This stage develops the model's capacity to execute structured reasoning under diverse analytical strategies. Using the Multi-perspective Reasoning Dataset (Section~\ref{sec:reasoning-dataset}), we train the model to generate Chain-of-Thought (CoT) rationales before producing final extractions. Formally, for each instance $I = \langle x,\, S,\, \tau_i,\, \textsc{CoT}_i,\, y_i \rangle$, we optimize:

\begin{equation}
\mathcal{L}_{\text{CoT}} = -\sum_{i=1}^{N} \sum_{p=1}^{P} \mathbb{E}_{\tau^p \sim \mathcal{T}}\log P(CoT_i^p | \mathcal{M}(x_i, S, \tau^p))
\end{equation}
\begin{equation}
\mathcal{L}_{\text{struct}} = -\sum_{i=1}^{N} \sum_{p=1}^{P} \mathbb{E}_{\tau^p \sim \mathcal{T}}\log P(y_i^p | \mathcal{M}(x_i, S, \tau^p))
\end{equation}

 The base model from Section~\ref{subsubsec1} is initialized with strategy-aware modifications. \textbf{Strategy Prefix Injection}: Prepend strategy descriptors $\tau_i$ to the input sequence. \textbf{Dynamic Decoding}: Introduce special tokens  \texttt{<think>} and \texttt{</think>} to segment reasoning steps and terminate CoT generation. \textbf{Multi-Task Objectives}:
\begin{equation}
  \mathcal{L} = \lambda_{\text{CoT}} \mathcal{L}_{\text{CoT}} + \lambda_{\text{struct}} \mathcal{L}_{\text{struct}}
\end{equation}

 To prevent strategy overfitting, we employ a regularization techniques:
\textit{Strategy Hiding}. We introduce an additional 10\% of training examples that instruct the model to skip all reasoning steps, emitting empty reasoning content between the markers \texttt{<think>} and \texttt{</think>}, and exclude the $\mathcal{L}_{\text{CoT}}$ for these samples during training.

The model dynamically adapts its reasoning path to novel strategy instructions while maintaining output validity against the unified schema.
\subsubsection{Reinforcement Learning for Alignment }\label{subsubsec3}
Explicitly specifying an optimal reasoning strategy for each individual instance during inference is often impractical: enumerating all possible strategies incurs prohibitive time overhead, while manually crafted rules struggle to adapt to the dynamic context-dependent nature of optimal strategies across diverse examples. To address this, we embed the selection of reasoning strategies as an intrinsic capability within the large language model itself. Given an input question $x$ and schema $S$, the model eschews explicit strategy prompting and instead performs end-to-end processing: internally handling strategy selection, reasoning , information extraction. To achieve this adaptive activation of optimal reasoning paths, we construct a multi-grained reward function and train the model using reinforcement learning.

We design a multi-granularity reward function to verify the correctness of prediction results and analyze the quality of the process.

\textbf{Result Reward \(R_{\text{result}}\)}.
We decompose the Result into category-level and argument-level correctness, re-weight them by $\alpha\!>\!\beta$ to reflect their unequal learning difficulty, and take the weighted harmonic mean to encourage joint optimization:
\begin{equation}
R_{\text{result}} = \frac{2\,\alpha\beta\,
      \mathbb{I}\!\left[y_{\text{c}} = y_{\text{c}}^{*}\right]\,
      \mathbb{I}\!\left[y_{\text{a}} = y_{\text{a}}^{*}\right]}
      {\alpha\,\mathbb{I}\!\left[y_{\text{c}} = y_{\text{c}}^{*}\right] 
     + \beta\,\mathbb{I}\!\left[y_{\text{a}} = y_{\text{a}}^{*}\right]},
\quad \alpha>\beta>0.
\end{equation}
Here $y_{\text{c}},y_{\text{a}}$ are predicted category and argument, $y^*$ are gold labels, $\mathbb{I}[\cdot]$ is the indicator function, and $\alpha>\beta$ encode the prior that category prediction is easier than argument extraction.

\textbf{Process Reward \(R_{\text{process}}\)}
The process quality is evaluated by:
\begin{equation}
R_{\text{process}} = \text{Faithfulness}(x,S,\tau) 
\end{equation}
where Faithfulness verifies whether the reasoning strictly follows the schema $S$, effectively leverages the information from $x$, and demonstrates a sound reasoning strategy $\tau$. The overall reward is a convex combination of the two components:
\begin{equation}
R = \lambda_{1}\,R_{\text{result}} + \lambda_{2}\,R_{\text{process}},
\quad \lambda_{1}+\lambda_{2}=1.
\end{equation}

We adopt a \textbf{rule-based Group Relative Policy Optimization (GRPO)} procedure. Initialize the policy with the SFT-Guidance model from section ~\ref{subsubsec2}, denoted $\pi_{\theta_{0}}$. For each training instance $(x,S,Y)$, sample $G$ complete reasoning chains from the old policy $\pi_{\theta_{\mathrm{old}}}$.
Compute per-chain rewards to update model.

\section{Experimental Setup}\label{sec4}
\subsection{Datasets}

\textbf{Supervised-tuning Datasets.}
We assemble a multi-task collection of human-annotated IE corpora that span three fundamental extraction paradigms (NER, RE and EE), and cover diverse domains including newswire, finance, biomedicine, and general science.
All corpora are re-standardized under a unified representation of schema.

\textbf{Zero-Shot Datasets.}
To rigorously assess the generalization capacity of our model under unseen-schema and cross-domain conditions, we adopt a strict zero-shot protocol.
Following the recommendation of YAYI-UIE~\citep{DBLP:journals/corr/abs-2312-15548}, we deliberately select evaluation sets that \emph{do not} overlap with the training corpora in either domain or schema.
These corpora collectively traverse literature, music, legal discourse, social media, and other domains \emph{never} exposed to the model during training, ensuring that the ensuing zero-shot results faithfully reflect the model’s adaptability to novel extraction patterns and unseen ontologies.
Comprehensive data statistics are reported in Table~\ref{tab:dataset_stats}.

\begin{table}[h]
\caption{Statistical data of IE datasets. \S ~marks zero-shot sets excluded from training. For datasets lacking an open test split, results are reported on the evaluation set. For EE tasks, Schemas are formatted as \emph{event (argument)} counts.}
\label{tab:dataset_stats}
\centering
\begin{tabular}{@{}c|c|c|c|c|c|c@{}}
\toprule
Task & Dataset & Domain & Schemas & Train & Val & Test\\
\midrule
 & CoNLL2003\citep{tjong-kim-sang-de-meulder-2003-introduction} & News & 4 & 12,613 & 3,070 & 3,184 \\
  & AnatEM\citep{10.1093/bioinformatics/btt580} & Biomedical & 1 & 5,667 & 2,081 & 3,758 \\
NER  & mit-restaurant\citep{6639301} & Social Media & 8 & 7,658 & -- & 1,520 \\
  & CrossNER\_Politics~\S\citep{Liu_Xu_Yu_Dai_Ji_Cahyawijaya_Madotto_Fung_2021} &  Political & 9 & -- & -- & 650 \\
  & CrossNER\_Music~\S\citep{Liu_Xu_Yu_Dai_Ji_Cahyawijaya_Madotto_Fung_2021} & Musical & 13 & -- & -- & 465 \\
\midrule
 & ADE Corpus\citep{Gurulingappa2012} & Biomedical & 1 & 3,416 & 427 & 428 \\
  & SciERC\citep{luan-etal-2018-multi} & Scientific & 7 & 1,366 & 187 & 397 \\
RE  & CoNLL2004\citep{carreras-marquez-2004-introduction} & News & 5 & 922 & 231 & 288 \\
  & KBP37\citep{zhang2015relationclassificationrecurrentneural} & News & 18 & 15,911 & 1,723 & 3,405 \\
  &  IPRE~\S\citep{10.1007/978-3-030-32236-6_9} &  General & 35 & -- & -- & 3,340 \\
\midrule
 & CASIE\citep{Satyapanich_Ferraro_Finin_2020} & Cybersecurity & 5(26) & 3,732 & 777 & 1,492 \\
  & PHEE\citep{sun-etal-2022-phee} & Biomedical & 2(16) & 2,897 & 960 & 968 \\
EE  & DUEE1.0\citep{10.1007/978-3-030-60457-8_44} & News & 65(217) & 11,908 & 1,492 & -- \\
  & DUEE-fin\citep{10.1007/978-3-031-17120-8_14} & Finance & 13(91) & 7,015 & 1,171 & -- \\
  & CrudeOilNews~\S\citep{lee-etal-2022-crudeoilnews} & Oil News & 18(104) & -- & -- & 356 \\
  & FewFC~\S\citep{Zhou_Chen_Zhao_Wu_Xu_Li_2021} &  Finance & 5(29) & -- & -- & 2,879 \\
\botrule
\end{tabular}
\end{table}

\subsection{Evaluation Metrics}
We adopt Micro-F1 as the primary evaluation metric. For NER task, models must accurately identify entity boundaries and their corresponding types. For RE task, models must precisely identify the subject and object entities involved in a relation and determine the specific relation type. In the case of EE task, we evaluate model performance on two subtasks: event detection and event argument extraction. Event detection requires the model to correctly identify the event type and its trigger span. Argument extraction requires the model to accurately identify predefined event arguments and their corresponding spans.
\subsection{Baselines}
To validate the effectiveness of the proposed MR-UIE model, we select two categories of baseline models for comparison: the first comprises advanced general large language models, which include Baichuan2-13B-Chat\citep{DBLP:journals/corr/abs-2309-10305}, LLaMA3-8B-Instruct\citep{DBLP:journals/corr/abs-2407-21783}, Qwen3-8B-Instruct\citep{yang2025qwen3technicalreport} and GPT-4\citep{openai2024gpt4technicalreport}. The second consists of state-of-the-art UIE-specific fine-tuned models:
\begin{itemize}
    \item UIE\citep{DBLP:conf/acl/0001LDXLHSW22} unifies text-to-structure generation for extraction tasks with schema-based prompts, leveraging the encoder-decoder power of generation model.
    \item InstructUIE\citep{DBLP:journals/corr/abs-2304-08085} unifies all information extraction tasks into a single, generalizable generative framework via multi-task instruction tuning.
    \item YAYI-UIE\citep{DBLP:journals/corr/abs-2312-15548} injects chat capabilities into universal information extraction through instruction tuning atop the open-source Baichuan2-13B.
\end{itemize}
Evaluations conducted on standard benchmark datasets (listed in Table~\ref{tab:dataset_stats}) demonstrate the generalization capability and superiority of MR-UIE.
\subsection{Implementation Details}
For Datasets, our training data are sampled from the corpus released with the IEPile paper.  
Because the open-source IEPile collection contains roughly two million training instances, the cost of generating multi-perspective reasoning annotations for the entire corpus would be prohibitive in both time and computational resources.  
We therefore first sub-sample eleven datasets that cover distinct domains.  
Subsequently, we randomly exclude instances with empty label sets, while retaining two categories of data: (i) all instances bearing at least one informative label; and (ii) a randomly selected 40\% of instances with empty label sets. This approach ensures a balanced yet computationally tractable training distribution.
We employ Deepseek-R1 to generate 15 reasoning guides for each training example excluding zero-shot scenarios. We then use Deepseek-R1 to produce reasoning processes and extract answers for 5 of these guides selected via algorithm ~\ref{alg:build-dataset-v1}, and add a feature field ``level'' indicating the count of successful correct-answer guidances. The SFT phase comprises two steps: first, we perform base fine-tuning without reasoning guides or reasoning modes using Qwen3-8B; second, we conduct multi-perspective reasoning SFT using the first-stage model, training exclusively on data with $\textit{level} > 2$. Both stages use a learning rate of $3 \times 10^{-5}$, train for 5 epochs with batch size 128, set maximum sequence length to 8,192, and utilize 8 A100 GPUs. 
\begin{figure*}[htbp]
    \centering
    \includegraphics[width=\linewidth, scale=5]{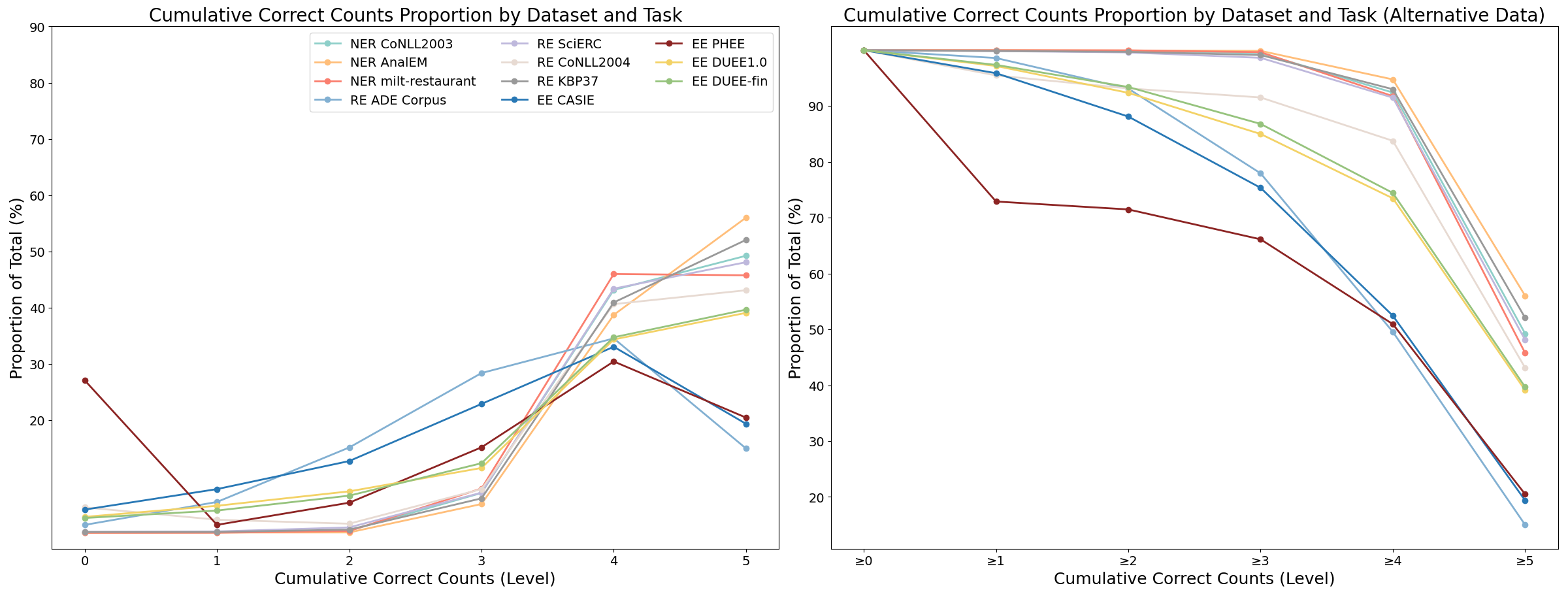}
    \caption{The distribution of the number of examples that successfully guided the correct answers in the training data construction phase for the NER, RE, and EE tasks under different strategies. The horizontal axis represents the number of times each example was guided to produce the correct answer, while the vertical axis shows the proportion of the corresponding examples in the total sample.}
    \label{fig:counts}
\end{figure*}

During the reinforcement learning phase, we leverage difficult samples to enhance the upper bound of model capabilities, while deliberately avoiding constraints on specific reasoning patterns. This encourages the model to autonomously explore and adopt optimal reasoning strategies. The curves in the Figure~\ref{fig:counts} display the cumulative proportion of correct answers for different tasks and datasets under different guidance frequencies, reflecting the differences among various strategies in guiding the correct answers. Specifically, we select training samples from the SFT dataset with levels $\le$ 2. For each question, the policy model generates 8 candidate outputs without any restriction on reasoning format, with a maximum output length of 2048. The training is conducted with a batch size of 128, a learning rate of 5e-7 for the policy model, and a KL coefficient of 0.01.

\section{Results and Analysis}\label{sec5}

\subsection{Overall Results in Supervised Learning}

We evaluate MR-UIE on eleven  supervised datasets; the numerical results are reported in Tables~\ref{tab:ner_results}--\ref{tab:ee_results}. 
Across five datasets, MR-UIE establishes new state-of-the-art performance in terms of Micro-F1, while on the remaining six datasets the gap to the respective SOTA never exceeds $3.2$ F1 points.

Table~\ref{tab:ner_results} summarizes the NER results. 
\begin{table}[h]
\caption{Over all supervision results on NER datasets. Results are reported as Micro-F1 scores.}
\label{tab:ner_results}
\centering
\begin{tabular}{@{}c|c|c|c|c|c@{}}
\toprule
Dataset & UIE & InstrucUIE & YAYI-UIE & IEPILE & MR-UIE \\
\midrule
CoNLL2003 & 92.99 & 92.94 & 96.77 & 92.98 & 92.61 \\
AnatEM & 77.68 & 90.89 & 76.54 & 86.90 & 87.21 \\
mit-restaurant & 81.67 & 82.55 & 79.38 & 81.3 & 83.50 \\
\botrule
\end{tabular}
\end{table}
On \texttt{mit-restaurant}, which requires the recognition of eight colloquial entity types characterized by ambiguous boundaries and fine-grained category distinctions, MR-UIE significantly outperforms all compared systems, demonstrating its superior capability in handling fine-grained classification. 
Conversely, \texttt{AnatEM} defines a single biomedical entity type, but the corpus contains dense domain-specific terminology and exhibits highly specialized contexts. 

Despite its restricted training size, MR-UIE ranks first, thereby confirming the model's generalizability to low-resource medical texts.

Results for RE and EE are presented in Tables~\ref{tab:re_results} and \ref{tab:ee_results}, respectively.
In RE, \texttt{KBP37} comprises eighteen relation categories with a highly skewed distribution; MR-UIE achieves the best overall scores. 
On the \texttt{ADE Corpus}, which contains only the single relation \textit{drug--adverse effect} and consists of short, noisy sentences, MR-UIE attains the second-best result.

In EE, \texttt{DUEE-fin} encompasses thirteen financial event types and ninety-one argument roles, featuring nested arguments and severe role overlap. 
MR-UIE ranks first on the argument extraction sub-task. 
The \texttt{PHEE} benchmark defines two biomedical event types and sixteen specialized roles, where argument spans are long and semantically ambiguous. 
MR-UIE secures the second position, further corroborating its stability in low-resource medical scenarios.

\begin{table}[h]
\caption{Over all supervision results on RE datasets. Results are reported as Micro-F1 scores.}
\label{tab:re_results}
\centering
\begin{tabular}{@{}c|c|c|c|c|c@{}}
\toprule
Dataset & UIE & InstrucUIE & YAYI-UIE & IEPILE & MR-UIE \\
\midrule
ADE Corpus & — & 82.31 & 84.14 & 85.87 & 80.12 \\
SciERC & 36.53 & 45.15 & 40.94 & 44.58 & 45.93 \\
CoNLL2004 & 75.00 & 78.48 & 79.73 & 73.71 & 73.55 \\
KBP37 & — & 36.14 & 59.35 & 61.49 & 63.48 \\
\botrule
\end{tabular}
\end{table}

\begin{table}[h]
\caption{Over all supervision results on EE datasets. Results are reported as Micro-F1 scores. We report event extraction results on two subtasks: trigger detection and argument extraction.}
\label{tab:ee_results}
\centering
\begin{tabular}{@{}c|c|c|c|c|c|c@{}}
\toprule
Task & Dataset & UIE & InstrucUIE & YAYI-UIE & IEPILE & MR-UIE \\
\midrule
 & CASIE & 69.33 & 67.8 & 63 & 61.27 & 69.51 \\
Trigger & PHEE & 64.77 & 70.14 & 63 & 68.52 & 71.02 \\
 & DUEE1.0 & 82.18 & — & 85 & 84.01 & 85.61 \\
 & DUEE-fin & 84.53 & — & 82.5 & 79 & 80.77 \\
\midrule
 & CASIE & 61.3 & 63.53 & 64.23 & 56.78 & 65.49 \\
Argument & PHEE & 63.7 & 62.91 & 77.19 & 71.33 & 73.0 \\
 & DUEE1.0 & 70.68 & — & 78.08 & 73.79 & 73.23 \\
 & DUEE-fin & 75.73 & — & 70.02 & 73.08 & 78.33 \\
\botrule
\end{tabular}
\end{table}

\subsection{Zero-shot performance}
In Table~\ref{tab:model_performance}, we report the zero-shot performance of MR-UIE across three tasks and two languages (English and Chinese, IPRE and FewFC are Chinese datasets). Overall, MR-UIE achieves top-two results on most tasks and outperforms GPT-4 in five experimental settings. Notably, it sets a new SOTA result for zero-shot NER on the CrossNER dataset. We attribute this success primarily to two key innovations: Unified Schema Representation: This strategy abstracts complex information extraction tasks into a schema mapping process. By standardizing the task format, it allows large models to focus solely on semantic understanding without needing structural adaptations for different tasks. Consequently, capabilities learned during supervised fine-tuning transfer readily to zero-shot scenarios. Adaptive CoT Reasoning: Leveraging query and schema features, this process dynamically decomposes problems into finer-grained sub-steps. For datasets like IPRE with numerous schemas, this approach is particularly effective, helping the model discern nuanced distinctions between easily confusable elements, thereby enhancing extraction accuracy. 
\begin{table}[h]
\centering
\caption{Zero-shot performance on IE task(NER, RE and EE). We report event extraction results on two subtasks: trigger detection and argument extraction.}
\label{tab:model_performance}
\begin{tabular}{@{}l c c c c c c@{}}
\toprule
 & \multicolumn{1}{c}{NER} & \multicolumn{1}{c}{RE} & \multicolumn{2}{c}{EE(Trigger)} & \multicolumn{2}{c}{EE(Argument)} \\
\cmidrule(lr){2-2} \cmidrule(lr){3-3} \cmidrule(lr){4-5} \cmidrule(lr){6-7}
Model & CrossNER & IPRE & CrudeOilNews & FewFC & CrudeOilNews & FewFC \\
\midrule
Baichuan2 & 38.93 & 1.45 & 0 & 11.82 & 0.48 & 6.91 \\
GPT-4 & 58.49 & 18.15 & 26.13 & 74.25 & 17.25 & 48.05 \\
LLaMA3 & 26.90 & 0.21 & 0.45 & 0.56 & 1.09 & 4.45 \\
Qwen3 & 29.40 & 0.37 & 2.47 & 5.21 & 1.22 & 9.33 \\
InstructUIE & 49.36 & -- & 23.26 & -- & 21.78 & -- \\
YAYI-UIE & 50.39 & 22.97 & 12.45 & 81.28 & 19.74 & 63.06 \\
IEPILE & 56.50 & 28.55 & 33.87 & 70.10 & 18.47 & 43.26 \\
MR-UIE & 62.21 & 23.31 & 21.22 & 83.80 & 13.51 & 56.00 \\
\botrule
\end{tabular}
\end{table}
\subsection{Ablation Study}
We evaluate the individual contribution of each core component by constructing three strictly-controlled variants of MR-UIE. All hyper-parameters, training data, and decoding strategies remain identical across variants. We design MR-UIE as a \emph{strictly sequential} pipeline for three complementary reasons. The ablation modules are detailed in Table ~\ref{tab:ablation_variants}.
\begin{table}[h]
\centering
\small
\setlength{\tabcolsep}{4pt}
\caption{Ablation variants of MR-UIE.}
\label{tab:ablation_variants}
\begin{tabular}{@{}lccc@{}}
\toprule
Variant & Unified Schema & Multi-Perspective SFT & RL Alignment \\ \midrule
MR-UIE (Full)   & \checkmark & \checkmark & \checkmark \\
w/o Schema      &  & \checkmark & \checkmark \\
w/o SFT     & \checkmark &  & \checkmark \\
w/o RL          & \checkmark & \checkmark &  \\ \bottomrule
\end{tabular}
\end{table}
\textbf{Unified Schema} first clamps the model to a \emph{closed output grammar}. Without this clamp, later multi-perspective or RL stages would waste capacity learning to recover from illegal formats; serialisation prevents this waste and ensures every subsequent gradient is spent on \emph{reasoning quality} rather than \emph{syntax repair}. During ablation experiments, we adopted IEPILE's schema representation. Specifically, NER, RE, and EE each have independent input and output structures. For the input, we reused MR-UIE's original reasoning strategy guidance. For the output, we retained the original reasoning process while only transforming the answer format. Furthermore, to adapt to the output formats of different subtasks, we designed the reinforcement learning reward function as three independent scenarios. Table~\ref{tab:ablation} reports Micro-F1 scores on representative datasets.
In the second row of Table~\ref{tab:ablation}, ablating this stage reduced F1 scores across four subtasks by 0.59, 0.42, 0.57 and 0.72 respectively.

\textbf{Multi-perspective reasoning data} injects distinct strategies per instance, yielding a \emph{beam of CoT rationales} for the same gold structure. When conducting the ablation study for this stage, we removed multi-perspective reasoning training and disabled the reasoning thinking mode. The second row in Table~7 shows that the multi-stage strategy is crucial to our architecture; its removal caused F1 decreases across all subtasks (maximum drop: 2.48), demonstrating the necessity of explicit multi-perspective fine-tuning.
\begin{table}[h]
\centering
\small
\caption{Micro-F1 scores of ablation variants.}
\label{tab:ablation}
\begin{tabular}{l>{\centering\arraybackslash}p{1.6cm}>{\centering\arraybackslash}p{1.6cm}>{\centering\arraybackslash}p{1.6cm}>{\centering\arraybackslash}p{1.6cm}}
\toprule
 & CoNLL2003 & SciERC & \multicolumn{2}{c}{CASIE} \\ \cmidrule(lr){4-5}
Variant                 & (NER)     & (RE)   & Trigger & Argument \\ \midrule
MR-UIE (Full)    & 92.61     & 45.93  & 69.51   & 65.49    \\
w/o Schema       & 92.02     & 45.51  & 68.94   & 64.77    \\
w/o SFT      & 90.13     & 44.12  & 67.22   & 63.88    \\
w/o RL           & 91.55     & 44.70  & 67.90   & 62.11   \\ \bottomrule
\end{tabular}
\end{table}
\textbf{RL Alignment} reframes strategy choice as a \emph{latent variable} inside the policy. GRPO rewards both \emph{result accuracy} and \emph{process faithfulness}, enabling the model to \emph{internally decide} which strategy to follow for any unseen $(x,S)$. Both variants outperform the single-template baseline, confirming that the RL stage has successfully \emph{internalised} the diverse strategies rather than memorised them.


\subsection{Multi-Perspective Reasoning Effectiveness}
\begin{figure*}[htbp]
    \centering
    \includegraphics[width=\linewidth, scale=5]{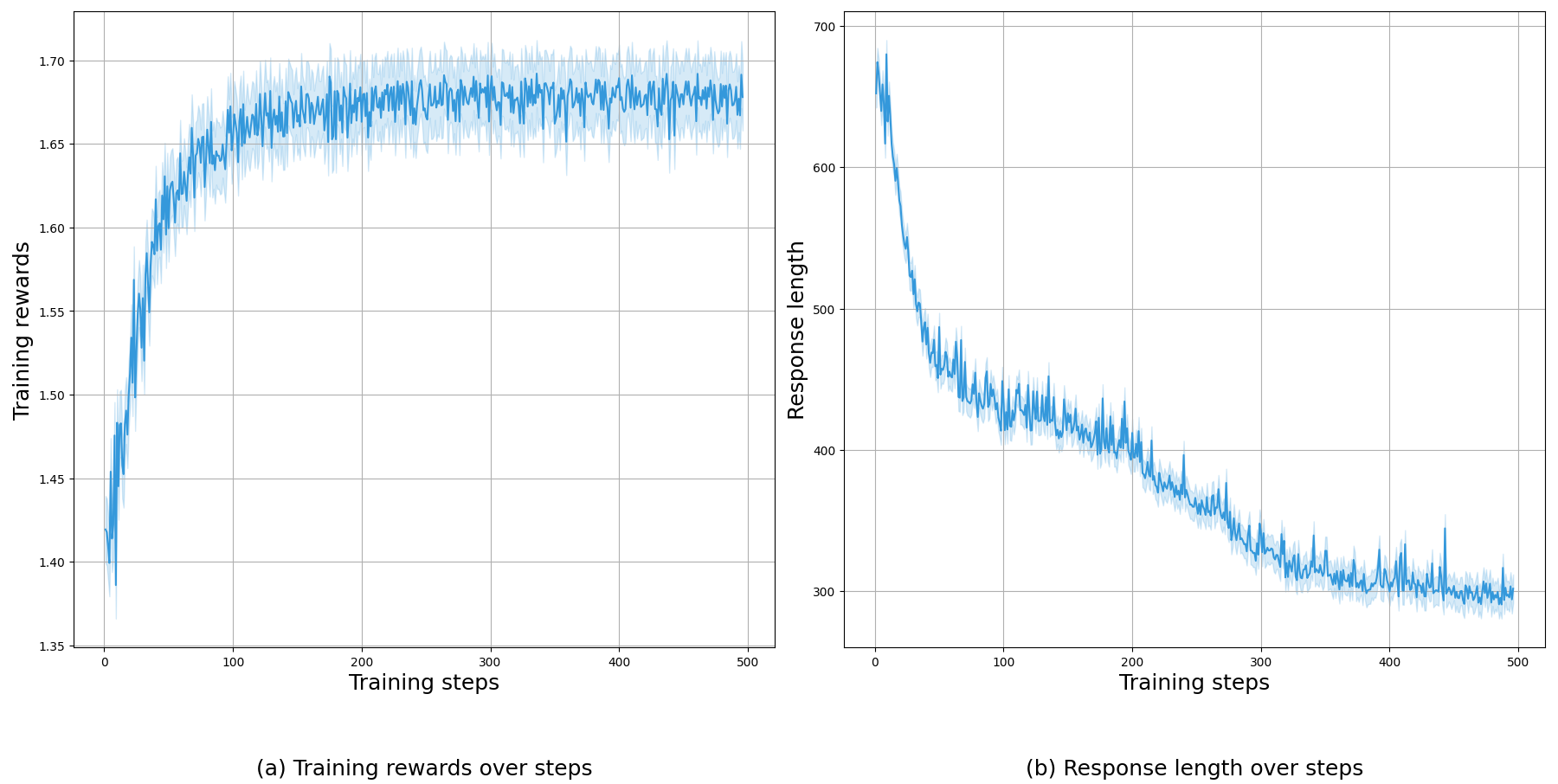}
    \caption{Reinforcement Learning Training Dynamics. The (a) panel illustrates the correlation between training steps and average reward value. The (b) panel examines the relationship between training steps and response length.}
    \label{fig:counts_reward}
\end{figure*}

This study systematically evaluates a baseline approach without Chain-of-Thought (CoT) alongside four distinct CoT strategies across two core information extraction tasks: event extraction (specifically trigger detection on the DuEE-fin dataset) and relation extraction (using the conll04 dataset), with performance quantified by F1 scores. The \textit{No Strategy} approach directly outputs final predictions without explicit reasoning steps. In contrast, \textit{Free Strategy} permits models to autonomously generate reasoning paths before producing outputs, absent external guidance. The \textit{Random Selection} method samples reasoning templates arbitrarily from a predefined strategy repository derived from training data, while \textit{Relevance Selection} identifies optimal templates through schema and query similarity matching against the same repository. Finally, \textit{Dynamic Strategy (RL)} enables models to freely generate reasoning pathways during "think" phases, with these capabilities refined through reinforcement learning to better align reasoning processes with task objectives. 
\begin{figure*}[htbp]
    \centering
    \includegraphics[width=\linewidth, scale=5]{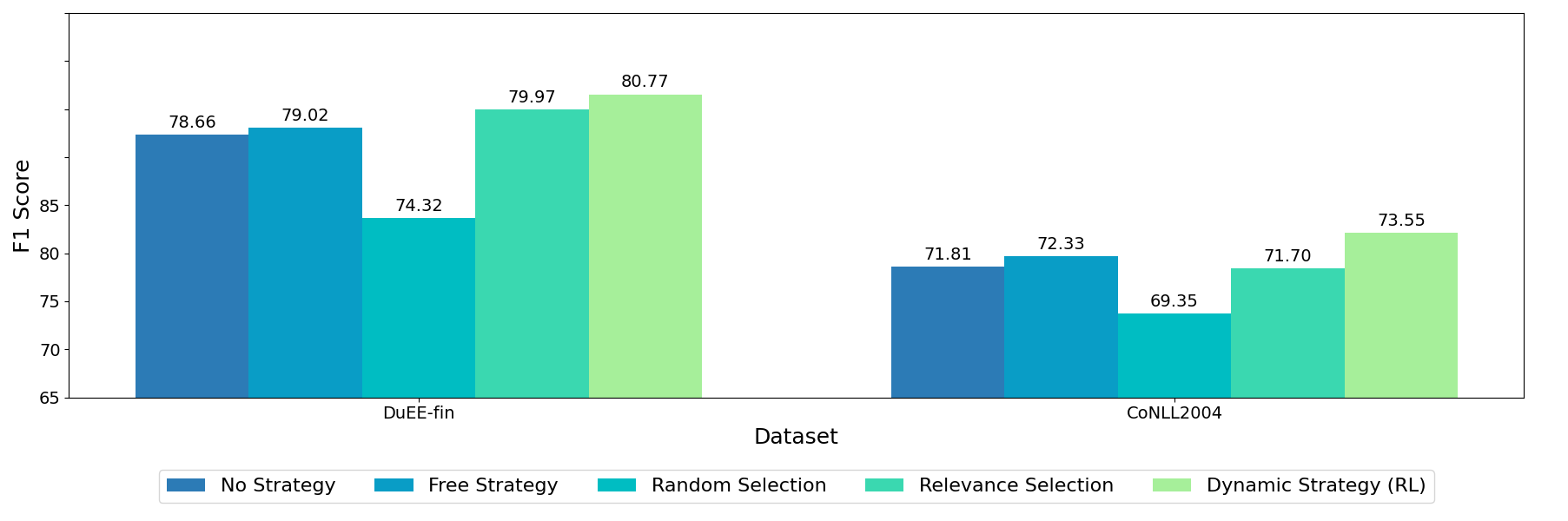}
    \caption{Analyzing the Impact of Different Chain-of-Thought Approaches on Model F1 Scores: A Comparative Study.}
    \label{fig:Figure_cot}
\end{figure*}
As revealed in Figure~\ref{fig:Figure_cot}, significant performance variations emerge across strategies and tasks. The demonstrable superiority of appropriate CoT strategies over the \textit{No Strategy} baseline validates the efficacy of explicit reasoning guidance in information extraction. Notably, \textit{Random Selection} substantially underperforms the baseline on both tasks, indicating that indiscriminate template application actively degrades performance by introducing misleading reasoning patterns. Task-dependent characteristics significantly influence outcomes: \textit{Relevance Selection} proves effective for event extraction yet underperforms in relation extraction, highlighting differential sensitivity to reasoning frameworks. The optimal performance of \textit{Dynamic Strategy (RL)} across both tasks underscores the advantage of adaptive, input-sensitive reasoning generation over static templates. By leveraging reinforcement learning to dynamically construct contextually relevant inference paths rather than relying on predetermined templates, this approach achieves superior flexibility in handling diverse semantic structures and query complexities, ultimately yielding more precise information capture.
\section{Conclusion}
Our paper introduce MR-UIE, a novel framework for universal information extraction. 
The MR-UIE model significantly advances the capabilities of LLMs in UIE. By fundamentally shifting the paradigm from treating LLMs as passive extractors to training them as adaptive reasoners, our approach explicitly models the complex reasoning processes required for structured output scenarios. This is achieved through the integration of multi-perspective reasoning with reinforcement learning optimization. Key to this success is our unified schema representation, which enables consistent handling of diverse elements like entities, relations, and events across tasks. The generation of multiple reasoning trajectories using varied strategies captures complementary inferential perspectives, while path-aware supervision during adaptive reasoning model training ensures the model learns robust reasoning procedures beyond merely predicting final answers. The application of RL, guided by a composite reward function evaluating correctness, faithfulness, and efficiency, allows the model to internalize and optimize the most effective reasoning patterns. Consequently, MR-UIE consistently elevates extraction accuracy across multiple domains and benchmarks, demonstrating superior performance compared to state-of-the-art methods. Crucially, the incorporation of explicit multi-perspective reasoning significantly enhances the model's generalization ability, particularly in complex scenarios, underscoring the indispensable role of structured reasoning in achieving robust and interpretable information extraction. This work establishes that empowering LLMs to learn how to reason is essential for tackling the challenges of UIE.

\section{Future Work}
Although MR-UIE enhances generalization through multi-perspective reasoning, the current framework still faces limitations in dynamically integrating diverse reasoning paths and handling highly noisy textual contexts. Future work could explore noise-aware reasoning modules and contrastive path selection mechanisms, drawing inspiration from recent advances in noise-robust representation learning and contrastive prompt tuning methods \citep{song2020tcnn,ning2023ump,xiong2024dual,zhang2024data,zhang2025variational,huang2025learn}. Additionally, further refinement of the reinforcement learning reward design could better capture complex structured reasoning patterns. Extending the unified schema-based approach to multimodal settings \citep{tao2006human,tao2008bayesian,gao2015learning,han2015two,jiang2018deep,huang2025enhance} also presents a promising direction, enabling more robust joint understanding of textual and visual information.

\newpage
\bibliography{main}

\end{document}